\documentclass[journal]{IEEEtran}

\usepackage{cite}
\usepackage{amsmath,amssymb,amsfonts}
\usepackage[ruled,vlined]{algorithm2e}
\usepackage{graphicx}
\usepackage{textcomp}
\usepackage{xcolor}
\usepackage{booktabs}
\usepackage{tabularx}

\def\BibTeX{{\rm B\kern-.05em{\sc i\kern-.025em b}\kern-.08em
    T\kern-.1667em\lower.7ex\hbox{E}\kern-.125emX}}
\begin{document}

\title{UDA-Restoration: Bridging Unsupervised Domain Adaptation and Multi-Degradation Denoising for Enhanced AV Semantic Segmentation}

\author{Zainab Bibi and Mohsin Kamal%
\thanks{Zainab Bibi and Mohsin Kamal are with the School of Electrical Engineering and Computer Science (SEECS), National University of Sciences and Technology (NUST), Islamabad, Pakistan. 

Mohsin Kamal is also in the school of computing at the University of Nebraska-Lincoln, USA (e-mail: zbibi.msds23seecs@seecs.edu.pk; mkamal3@unl.edu).}%
}

\maketitle

\begin{abstract}
Autonomous vehicles rely heavily on vision-based perception systems for safe navigation and decision-making. Adverse weather conditions, sensor noise, moving artifacts and illumination changes, however, often degrade the image quality and hence reduce perception performance. This paper introduces a diffusion-guided unsupervised domain adaptation algorithm for autonomous driving environments, namely UDA-Restoration, which is designed to enhance and denoise images. The proposed method follows the paradigm of Noise-Space Domain Adaptation (Noise-DA), which is based on multi-source degradation learning and diffusion-guided adaptation that spans across synthetic and real-world degradations. The framework has been extensively tested and shown to restore image quality and to be robust to unseen degradation scenarios, such as snow, haze, sensor distortions, and more. The quantitative results indicate that Noise-DA outperforms the evaluated self-supervised and unsupervised baselines while matching the best supervised restoration performance. In addition, semantic segmentation experiments show that the performance of downstream perception tasks is directly improved by better image restoration. The outcomes show the promise of unsupervised domain adaptation based on diffusion as a powerful and scalable method for image restoration tasks in autonomous driving systems under various environments.
\end{abstract}


\section{Introduction}
\IEEEPARstart{C}{amera}-based perception systems are integral to many of the functions in autonomous vehicles (AVs) \cite{30Kumar2025VisionTechnologies}, including lane detection, obstacle avoidance, traffic sign recognition, and scene understanding. Recent research in deep learning has brought the ability to perceive the visual world to the level of high accuracy \cite{Zhang2023DeepLearningIQA}, but the accuracy is very susceptible to image quality. Images captured by a camera or sensor are often corrupted by weather conditions, lighting changes, sensor artifacts, distortion due to motion, and many other factors in real-world driving scenarios. Rain, fog, snow, motion blur, glare, low-light conditions, and sensor noise degrade scene visibility \cite{31Verma2025WeatherAffectedVision} thus affecting tasks in the downstream.
\begin{figure}[!t]
    \centering

    \includegraphics[width=0.9\columnwidth]{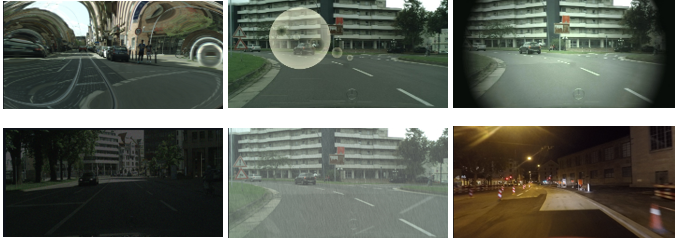}

    \caption{Examples of Image Degradations in Autonomous Driving Environments}
    \label{fig:fig1}
\end{figure}
Fig.~\ref{fig:fig1}. shows image degradations in autonomous driving situations such as rain, fog, snow, motion blur, low light, glare and sensor noise. These are degradations that affect the visibility of the scene and can have a negative impact on downstream perception tasks like object detection, lane detection, and semantic segmentation.

Thus, image restoration, denoising, has become a vital part of the autonomous driving pipelines \cite{43Hossain2025AdaptiveAerialRestoration}. In recent years, many deep learning-based image restoration models have been developed that have successfully restored images from paired clean-degraded image sets \cite{45Shi2026FIRSDE}, such as DnCNN, Restormer, and NAFNet. Gathering ideally matched training pairs, across various real-world driving scenarios, however, can be expensive and impractical. Restoration models are often supervised, and they tend to fail to generalize to other instances of degradation and other domain.

To decrease the reliance on paired supervision, self-supervised and unsupervised restoration methods like Noise2Noise, Noise2Void and Blind2Unblind have been introduced. Although these methods offer flexibility in training, the restoration ability is usually degraded in extreme real-world degradation. Furthermore, most previous work emphasizes only quality metrics like PSNR and SSIM, and how the restoration process affects the autonomous driving perception is under-researched. Synthetic-to-real domain adaptation and diffusion-based restoration have demonstrated good performance for synthetic-to-real restoration in the field of image restoration. Diffusion based restoration and domain adaptation techniques have been demonstrated to have promising performances in bridging the gap between synthetic and real degradations in the field of image restoration. Specifically, Noise-Space Domain Adaptation (Noise-DA) \cite{46Liao2025DenoisingAsAdaptation} aligns in the diffusion noise space, allowing for robust restoration without target domain paired supervision. This approach enhances the generalization of restoration and offers a viable approach for processing a variety of degradation situations found in actual driving conditions.

\begin{figure}[!t]
    \centering

    \includegraphics[width=0.9\columnwidth]{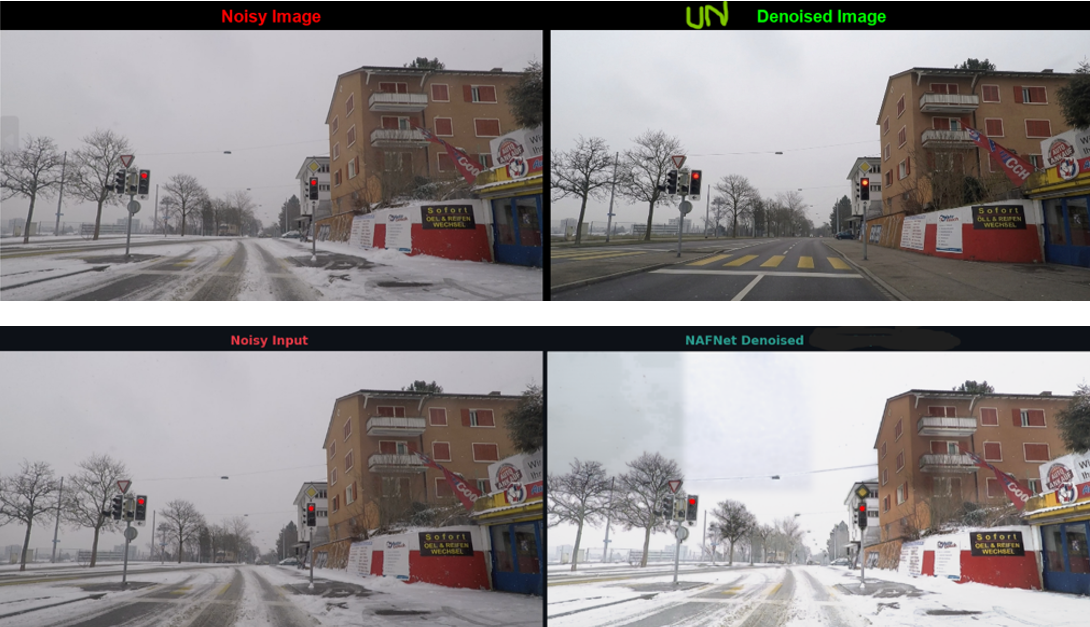}

    \caption{Qualitative Comparison Between Supervised and Unsupervised Restoration Methods}
    \label{fig:fig2}
\end{figure}

Fig.~\ref{fig:fig2}. shows the Qualitative comparison of the performance of the restoration on real-world degradations unseen in the test set, such as snow-covered scenes, and sensor-induced barrel distortions. The proposed Noise-DA framework performs well in the restoration task, especially when the domain shift happens with different degradation types, and the supervised NAFNet model shows limited generalization ability under domain shift.

Inspired by these advancements, this paper introduces UDA-Restoration, an image restoration method for autonomous driving that is based on diffusion. The proposed approach integrates supervised, self-supervised, and unsupervised restoration paradigms, and testifies their performance under different degradation situations. Moreover, qualitative image quality evaluation is not just defined by quantitative image quality metrics, but also by downstream semantic segmentation performance. A multi-source training strategy is used to enhance the performance against unseen degradations, and qualitative assessments of the snow and sensor-distortion cases are also conducted to show the advantages of diffusion-guided domain adaptation.

This work makes the following main contributions:

\begin{itemize}
    \item A unified image restoration framework is provided to assess various supervised, self-supervised and unsupervised image restoration methods under different degradation scenarios.

    \item A multi-source training diffusion-guided Noise-DA method is used to enhance cross-domain generalization and robustness in restoration.

    \item Extensive experiments are carried out in nine degradation categories to benchmark the restoration to be done in autonomous driving environments.

    \item The study explores the impact of image restoration on downstream semantic segmentation, using a set of segmentation architectures.

    \item Qualitative assessments on snow and sensor-distortion scenarios show that Noise-DA outperforms supervised restoration models in terms of its generalization ability.
\end{itemize}

\section{RELATED WORK}
This section reviews the literature most relevant to the proposed UDA-Restoration framework. The discussion covers image restoration and denoising, autonomous vehicle perception under adverse weather conditions, unsupervised domain adaptation for autonomous driving, and diffusion-based restoration and domain adaptation. Together, these studies establish the motivation for combining restoration, domain adaptation, and downstream perception evaluation within a unified framework.

\subsection{Image Restoration and Denoising}
In the context of autonomous driving under harsh conditions, image restoration holds an important position in enhancing the reliability of the driving system. In recent years, deep learning has made a huge impact on the ability to recover good image from degraded observations, which are corrupted by rain, fog, snow, motion blur, and sensor noise. However, restoring images captured under complex real-world conditions remains a challenging problem due to the diversity and unpredictability of environmental degradations.

Li et al. \cite{13Li2024WeatherGeneralSpecific} presented a unified weather restoration framework by disentangled representation learning, in which weather specific degradations and clean scene structures were learned separately as factors. Experimental results on Outdoor-Rain, RainDrop, and Snow100K datasets gave a PSNR of 25.31 dB, and preserved structural consistency under various weather conditions. The qualitative results showed that it was effective to remove the snow, fog, and rain artifacts and to keep the traffic boundaries and fine scene details.

To combat these compounding weather degradations, An et al. \cite{14An2025MoEDriven} proposed a Mixture-of-Experts (MoE) architecture which dynamically allocates image regions to the specialized branches for the restoration task. In terms of performance, their framework reached a peak PSNR of 1.2 dB and kept the real-time processing speed of 45 FPS. Likewise, Wang et al. \cite{15Wang2025MOERL} added reinforcement learning to an MoE restoration pipeline that enabled adaptive feature cleansing and resulted in a 0.94 SSIM with a computational complexity reduction of about 30\%.

Lin et al. \cite{16Lin2025MultiComponent} introduced a multi-component decomposition framework for mixed-weather restoration, which decomposed degraded images into various frequency components and then restored them using a transformer. Their approach outperformed traditional restoration techniques with an improvement in PSNR of 1.8 dB and showed better results in recovering high-frequency textures and edge structures. In a later work, Fang et al. \cite{17Fang2025WeatherCycle} investigated the fully unsupervised restoration based on color space decoupling and cycle-consistent learning, which resulted in a PSNR of 24.15 dB and SSIM of 0.89 on real-world driving datasets.

Recently, diffusion-based restoration methods have attracted much interest. Wen et al. \cite{24Wen2025AllInOneWeather} introduced an Adaptive Degradation-Aware Self-Prompting Model (ADSM), which adds CLIP-informed latent prompts to a diffusion restoration model. The extensive test results showed good PSNR and SSIM values on various weather degradation scenarios. The qualitative results revealed successful suppression of compound degradations such as rain, snow and fog, and good preservation of image details. 

Likewise, Cui et al. \cite{25Cui2025MultiScaleFeature} proposed a Multi-Scale Feature Modulation Network (MSFM), which integrates the frequency-domain features with the spatial-domain features to assist in autonomous vehicle perception image restoration. Their framework has improved the visibility of objects and the clarity of the scenes considerably in adverse environmental conditions.

\subsection{Autonomous Vehicle Perception Under Adverse Weather Conditions}
For the autonomous vehicle, the performance of downstream perception modules is seriously influenced by environmental degradations. Weather conditions can affect the quality of semantic segmentation, object detection, tracking, and scene understanding, all of which can have a negative impact on driving safety.

As weather is often considered as a critical component to the perception of autonomous vehicles, Zheng et al. \cite{26Zheng2023RobustPerception} introduced a robust perception framework for adverse weather conditions using generative data augmentation. The framework successfully incorporated realistic variations of weather conditions in the training process, which led to the significant enhancement of object detection and semantic segmentation performance on the KITTI, nuScenes, Cityscapes, and BDD100K datasets. Qualitative tests showed greater resistance to rain streaks, fog and low light conditions.

Liu et al. \cite{18Liu2025WeatherNormalization} proposed a two-stage perception framework to separate the weather normalization and semantic scene understanding. The framework implemented a generative weather enhancement module and then a frozen semantic segmentation network, enhancing ACDC with a 4.8\% mIoU improvement. Wang et al. \cite{21Wang2025TaskDrivenEnhancement} also explored task-driven image enhancement directly driven by segmentation loss functions. They can achieve an improvement of 6.3\% mIoU in semantic segmentation, proving that the quality of restoration should not be judged by image quality metrics alone.

DiffAD is a recent autonomous driving framework proposed by Wang et al. \cite{20Wang2025DiffAD} that incorporates probabilistic noise modelling into the Bird's-Eye View (BEV) perception pipelines. The framework enhanced the segmentation accuracy of BEVs by 3.1\% mIoU and tracking performance by 2.4\% AMOTA without degrading the navigation performance under severe environmental degradation.

\subsection{Unsupervised Domain Adaptation for Autonomous Driving}
Unsupervised Domain Adaptation (UDA) has proven to be a promising solution to the problem of mismatched distribution between labeled source domains and unlabeled target domains. Given that autonomous driving requires collecting pixel-level annotations under all weather conditions, which is impossible, UDA is especially significant in this type of driving.

One of the first UDA approaches for the weather domain was proposed by Erkent and Laugier \cite{27Erkent2020SemanticSegmentation} who used adversarial and statistical distance minimization methods to align feature distributions in clear and adverse weather states. Their method achieved a substantial gain under different weather conditions for semantic segmentation task without using target domain annotations.

Recent developments have stressed adaptation strategies that have been linked to weather. Shen et al. \cite{1Li2025WControlUDA} introduced W-ControlUDA, a diffusion-assisted domain adaptation framework that can create controllable weather from foundation diffusion models. Experimental results on GTA5, Cityscapes and ACDC datasets showed improvements of 4.3\% mIoU for foggy weather and 5.1\% mIoU for snowy weather. The framework maintained the integrity of the objects and avoided the structural distortions which are typical of oppositional translation models.
Pan et al. \cite{3Pan2025WeatherAware} proposed a framework for aggregation and adaptation based on weather conditions, which introduced weather-specific adapters and prototype-based contrastive learning to solve feature heterogeneity under different weather conditions. Their approach has made the Cityscapes-to-ACDC adaptation benchmark better by 4.6\% mIoU. In a similar fashion, Truong et al. \cite{4Truong2024CONDA} presented CONDA, a continual unsupervised domain adaptation framework for long-term autonomous driving deployment. CONDA achieved 18.4\% reduction in catastrophic forgetting and 3.2\% mIoU improvement in adaptation to new environments, by using an exponential moving average teacher model and pseudo-label refinement continuously.

Yu et al. \cite{6Yu2025ADLD} further investigated Inter-Coder Connected Latent Diffusion (ICCLD), a framework based on diffusion for latent adaptation. Adaptation evaluation for GTA5-to-Cityscapes showed a marginal gain of 2.7\% mIoU with the same geometric scene structures. Other adaptation studies have been conducted for night vision perception and aerial tracking \cite{8Ye2022NighttimeTracking, 9Wei2024NighttimeObjectTracking, 22Pan2025ObjectAwareUDA, 28Pan2024DynamicSmallObjects}, which further demonstrate the increasing relevance of domain adaptation under various demanding environments.

However, most UDA methods focus mainly on high-level semantic segmentation tasks, with limited works dedicated to image restoration and degradation removal.
\subsection{Diffusion-Based Restoration and Domain Adaptation}
In recent years, diffusion models have been shown to be a useful class of generative model for images, which can be useful in tasks such as image generation, restoration, and domain adaptation, because they can model complex data distributions progressively. Diffusion-based methods feature stable optimization and better structural preservation, which is not possible in the traditional adversarial approach.

Peng et al. \cite{10Peng2025DomainAdaptiveDiffusion} presented a progressive distribution alignment framework called Domain-Adaptive Diffusion for domain adaptation that operates via a progressive diffusion transition. Experimental results showed that this led to a 3.4\% mIoU gain for the Cityscapes and ACDC benchmarks and to the reduction of structural artifacts in the adaptation process. Ridley et al. \cite{11Ridley2025ControllingSemantics} also investigated semantic-preserving diffusion augmentation with ControlNet and LoRA-based adaptation modules, with the results showing a 15\% decrease in FID and a 2.1\% increase in segmentation accuracy.

Benbarka et al. \cite{12Benbarka2025SeeingThroughWeather} studied how to simulate controlled climate with generative diffusion models for robust perception in challenging weather conditions. Their framework enhanced the object tracking performance by 8.4\% and produced natural weather degradations without manual annotations.

Among diffusion-based adaptation approaches, Noise-Space Domain Adaptation (Noise-DA) proposed by Liao et al. \cite{7Liao2025DenoisingAdaptation} represents one of the most relevant studies to this work. Noise-DA is different from feature space/adaptation method and pixel space/adaptation method, because it directly aligns the domain in the diffusion noise space. It takes advantage of the diffusion-guided denoising objective along with channel shuffling and residual swapping contrastive learning mechanisms to avoid shortcut learning. Noise-DA’s performance was tested on SIDD and DND benchmarks, and it was found to be able to remove real-world sensor noise and environmental artifacts with a gain of around 1.15 dB PSNR. Qualitative results showed that the proposed framework preserved the details and fine textures better than conventional adaptation frameworks.

\section{PROPOSED METHODOLOGY}
This section presents the complete methodology of the proposed UDA-Restoration framework. The workflow combines multi-source degradation preparation, supervised and unsupervised image restoration, diffusion-guided Noise-Space Domain Adaptation, and downstream semantic segmentation evaluation. The overall processing sequence is first described, followed by the individual restoration, adaptation, and perception components.

\subsection{Overview of the Proposed Framework}
The proposed UDA-Restoration framework is designed to enhance the image quality and downstream perception performance in autonomous driving scenarios with various degradations. The framework is composed of diffusion-based unsupervised domain adaptation and multi-source noise learning for bridging the gap between synthetic and real-world image degradation.

\begin{figure}[!t]
    \centering
    \includegraphics[width=0.9\columnwidth]{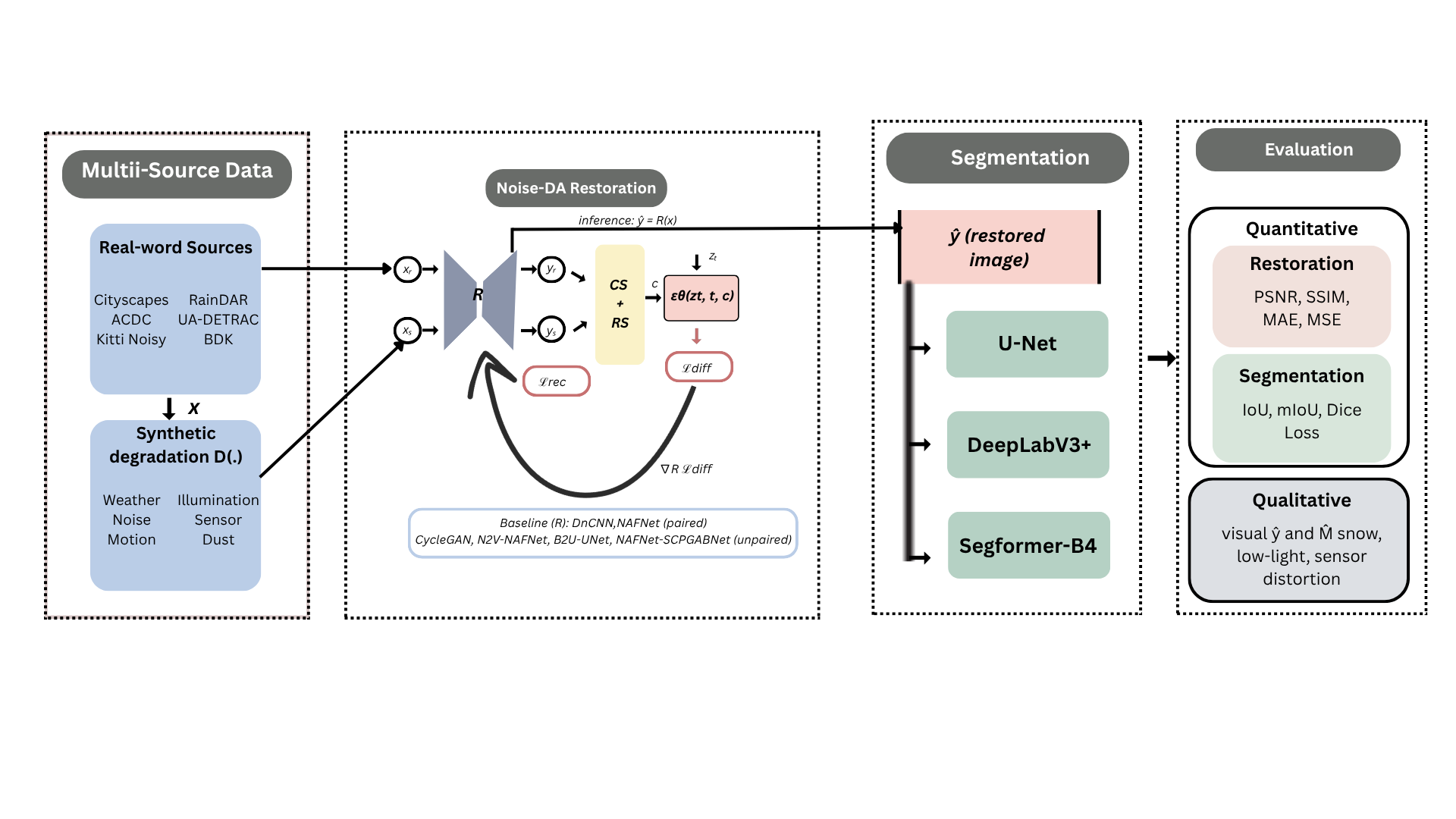}

    \caption{Overall Proposed UDA-Restoration Framework for Image Enhancement and Perception Evaluation in Autonomous Driving}
    \label{fig:fig3}
\end{figure}

As shown in Fig.~\ref{fig:fig3}, the framework consists of four key stages: (i) multi-source dataset preparation, (ii) image restoration by Noise-Space Domain Adaptation (Noise-DA), (iii) perception evaluation by semantic segmentation, and (iv) quantitative and qualitative performance evaluation.

In the first phase, several real-noise data sets, such as Cityscapes, ACDC, kitty, RainDAR, BDK and UA-DETRAC are integrated with degradation effects generated through synthetic methods, including Gaussian noise, motion blur, rain and snow, glare, low illumination and vignette. This allows the restoration model to be trained in multiple domains with good degradation representations. In the second stage, the degraded images are restored by directly applying the Noise-DA framework in the diffusion noise space. The restoration network and diffusion model are optimally coupled to produce similar degradations from synthetic and real images, to enhance restoration robustness and cross-domain generalization.

The restored images are then fed to various semantic segmentation architectures to understand the influence of image restoration on the perception. Lastly, quantitative indicators of restoration effectiveness, including PSNR, SSIM, IoU, and mIoU, and qualitative visual evaluation are used to evaluate the restoration effectiveness and generalization capability under real-world challenges.

The complete processing sequence of the proposed UDA-Restoration framework is summarized in Algorithm~\ref{alg:uda_restoration}.

\begin{algorithm*}[!t]
\caption{UDA-Restoration Framework for Image Restoration and Perception Evaluation}
\label{alg:uda_restoration}
\DontPrintSemicolon
\SetAlgoLined
\small

\KwIn{Multi-source clean and degraded driving images}
\KwOut{Restored images and restoration/perception evaluation results}

\BlankLine
\textbf{Stage 1: Data Preparation}\;
1. Collect images from Cityscapes, ACDC, KITTI Noisy, RainDAR, UA-DETRAC, and BDK\;
2. Generate synthetic degradations: Gaussian noise, motion blur, rain, snow, fog, glare, low illumination, vignette, and sensor distortion\;
3. Construct paired and unpaired training sets\;

\BlankLine
\textbf{Stage 2: Image Restoration}\;
4. Train supervised baselines using paired clean--degraded images\;
5. Train self-supervised and unsupervised models using unpaired degraded images\;
6. Apply Noise-DA to synthetic and real degraded samples\;

\BlankLine
\textbf{Stage 3: Noise-Space Domain Adaptation}\;
7. Restore synthetic and real degraded inputs using the shared network $R(\cdot)$\;
8. Condition the diffusion model on restored representations\;
9. Apply channel shuffling and residual swapping\;
10. Align synthetic and real degradation distributions in the diffusion noise space\;

\BlankLine
\textbf{Stage 4: Restoration Evaluation}\;
11. Evaluate restored images using PSNR and SSIM\;
12. Perform qualitative evaluation under seen and unseen degradations\;

\BlankLine
\textbf{Stage 5: Perception Evaluation}\;
13. Feed degraded and restored images to U-Net, DeepLabV3, and SegFormer\;
14. Evaluate segmentation using IoU, mIoU, and Dice score\;
15. Compare perception performance before and after restoration\;

\BlankLine
\Return{Restored images and quantitative/qualitative evaluation results}\;

\end{algorithm*}

\subsection{Image Denoising Framework}
Image denoising is the first step of the proposed UDA-Restoration framework, which is designed to restore clean visual information from degraded driving scenes. Adverse weather conditions, illumination changes, sensor noise, motion artifacts and optical distortions in autonomous driving environments affect image quality, consequently, negatively influence downstream perception tasks like semantic segmentation and scene understanding.

To assess restoration performance under varying degradation conditions, both paired (supervised) and unpaired (unsupervised) denoising methods were studied. Paired methods learn a direct restoration mapping from corresponding degraded and clean image pairs, whereas unpaired methods learn restoration without requiring matched clean ground-truth images. This setup enables a systematic comparison between conventional supervised restoration and domain-adaptive unsupervised approaches.
The degradation considered covers rain, fog, snow, gaussian noise, motion blur, low-light, glare, sensor distortions, dust and illumination variations. The datasets and restoration models used in this study are summarized in Table ~\ref{tab:tab1}.

\begin{table}[!t]
\caption{Datasets and Restoration Models Used in the Denoising Framework}
\label{tab:tab1}
\centering
\footnotesize
\begin{tabularx}{\columnwidth}{p{2.2cm} X X}
\toprule
\textbf{Category} & \textbf{Datasets} & \textbf{Models} \\
\midrule
Paired Denoising &
Cityscapes, ACDC (Augmented) &
DnCNN, NAFNet \\
Unpaired Denoising &
Cityscapes+Noise, KITTI Noisy, RainDAR, UA-DETRAC, ACDC, BDK &
CycleGAN, NAFNet-SCPGANet, Noise2Void-NAFNet, Blind2Unblind-UNet, Noise-DA \\
\bottomrule
\end{tabularx}
\end{table}

This study evaluates two denoising approaches, paired and unpaired, which are described in the following subsections.

\subsubsection{Supervised Denoising}
Paired denoising approaches train a direct mapping function between noisy and clean images based on noisy-clean image pairs. This paper contains paired denoising experiments on Cityscapes \cite{48Sowe2025PSPNetCityscapes} and ACDC \cite{47Sakaridis2025ACDC} images in the presence of synthetic degradations: rain, fog, gaussian noise, motion blur, low-light, glare, and sensor artifacts.

\begin{figure}[!t]
    \centering
    \includegraphics[width=0.9\columnwidth]{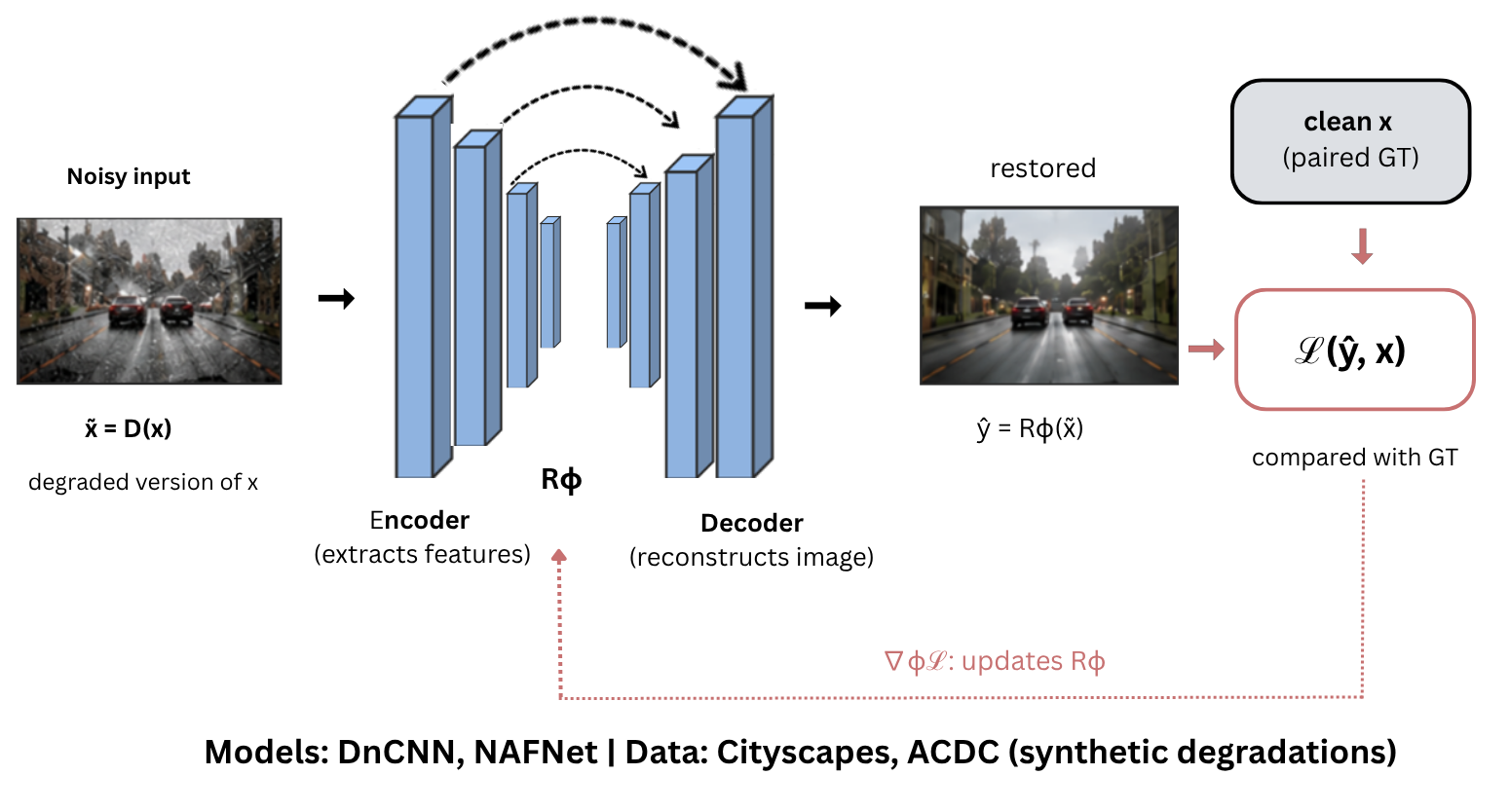}

    \caption{Supervised image denoising framework}
    \label{fig:fig4}
\end{figure}

Two supervised restoration models, DnCNN and NAFNet, were tested to explore their restoration capacity of recovering clean scene structure and maintaining the visual information for downstream perception tasks. Fig.~\ref{fig:fig4}, shows the supervised denoising pipeline. The driving images are also processed with synthetic degradation to create pairs of driving images for training, that allows for a direct mapping between the degraded images and the clean ones to be learnt by restoration models.

\subsubsection{Unsupervised Denoising}
It is usually difficult to get a pair of clean and degraded driving images in the real environment. Thus, unpaired denoising methods were studied to find out the restoration mappings without having any clean reference images.

Under various degradation conditions, several unsupervised restoration methods, such as CycleGAN, Noise2Void-NAFNet, Blind2Unblind-UNet, NAFNet-SCPGANet, and Noise-DA are studied. These methods use adversarial learning or self-supervision or diffusion-guided adaptation to recover the corrupted scenes.

\begin{figure}[!t]
    \centering
    \includegraphics[width=0.9\columnwidth]{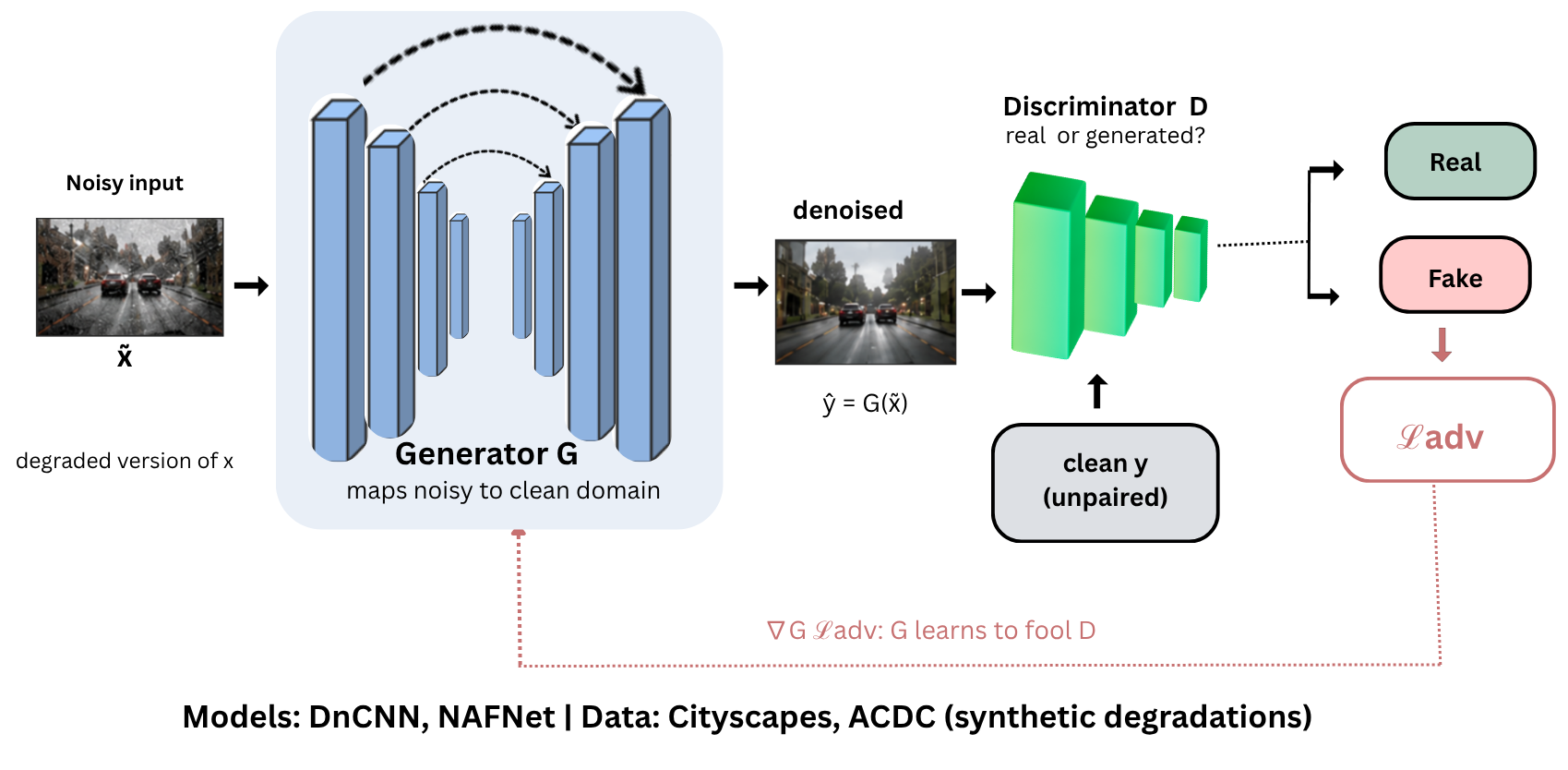}

    \caption{UnSupervised image denoising framework}
    \label{fig:fig5}
\end{figure}

Unsupervised denoising framework used in this study is shown in Fig.~\ref{fig:fig5}. Images with degraded data are restored under adversarial, self-supervised or domain adaptation learning strategies. Noise-DA showed the best generalization performance in unseen degradation categories and was thus chosen as a core method of noise restoration for further perception experiments.

\subsubsection{Selected Restoration Models}
For a comprehensive assessment as shown in Table ~\ref{tab:tab2}, a range of different learning paradigms were chosen supervised, adversarial, self-supervised and diffusion-based approaches. The selected models are widely used image restoration methods and are used to make a balanced comparison with various autonomous driving degradations.

\textbf{DnCNN:} A supervised restoration baseline based on CNN (DnCNN). \\
\textbf{NAFNet:} The NAFNet (Activation-free image restoration network) performs well in denoising, and it is a efficient architecture network. \\
\textbf{CycleGAN:} An image translation framework based on GANs that can be trained without paired data. \\
\textbf{Noise2Void-NAFNet:} Blind-spot learning and NAFNet based self-supervised restoration approach. \\
\textbf{Blind2Unblind-UNet:} Self-supervised denoising network based on U-Net model for image denoising. \\
\textbf{Noise-DA:} A framework for domain adaptation based on diffusion, with the restoration model that is the focus of investigations in this study. \\

\begin{table}[!t]
\caption{Characteristics of Restoration Models Used in This Study}
\label{tab:tab2}
\centering
\footnotesize
\setlength{\tabcolsep}{3pt}
\begin{tabularx}{\columnwidth}{p{0.25\columnwidth} X p{0.22\columnwidth}}
\toprule
\textbf{Model} & \textbf{Learning Paradigm} & \textbf{Supervision Type} \\
\midrule
DnCNN & CNN-based Denoising & Supervised \\
NAFNet & CNN-based Restoration & Supervised \\
CycleGAN & GAN-based Translation & Unsupervised \\
Noise2Void-NAFNet & Self-Supervised Denoising & Unsupervised \\
Blind2Unblind-UNet & Self-Supervised Restoration & Unsupervised \\
Noise-DA & Diffusion-Based Domain Adaptation & Unsupervised \\
\bottomrule
\end{tabularx}
\end{table}

\subsection{Noise-DA Based Unsupervised Domain Adaptation}
\begin{figure}[!t]
    \centering
    \includegraphics[width=0.65\columnwidth]{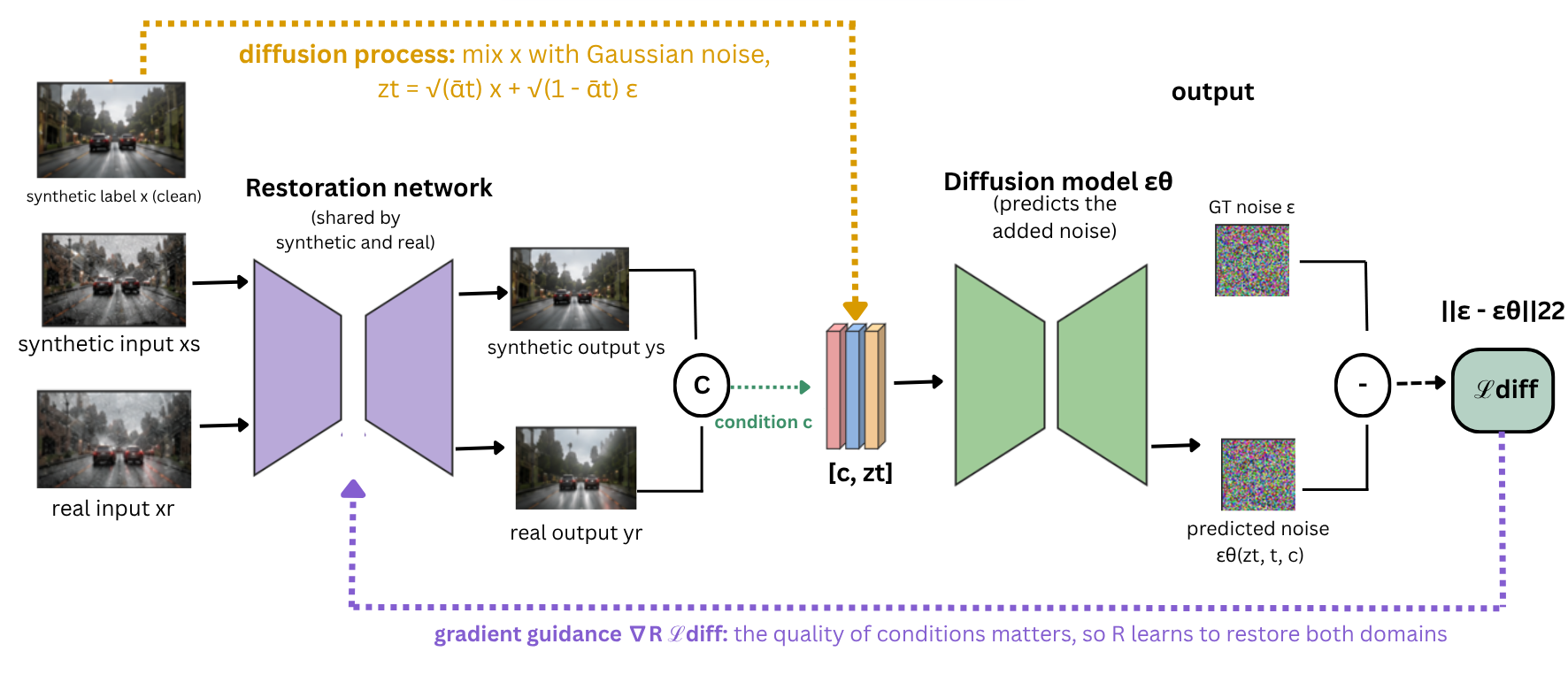}

    \caption{Architecture of the Noise-Space Domain Adaptation (Noise-DA) framework}
    \label{fig:fig6}
\end{figure}

In this study, Noise-Space Domain Adaptation (Noise-DA) \cite{46Liao2025DenoisingAsAdaptation} was found to have the best ability to generalize and was used as the primary restoration framework for this study. In contrast to traditional supervised denoising techniques, Noise-DA restores a noisy image by diffusion-based adaptation in the noise space, allowing for alignment of the distributions of noise degradation on the synthetic and real images.
The framework is shown in Fig.~\ref{fig:fig6}. works together by taking both the synthetic and real degraded images and using a common restoration network and diffusion model to find a common distribution of clean images. In addition, channel shuffling and residual swapping operations are used to boost domain invariance and boost the generalization to unseen domains, which improves the ability to recover from unseen degradation conditions.

A restoration network and diffusion model are both used to process the degraded image, where the former is used to reconstruct synthetic degraded images, and the latter is used to process real degraded images. Diffusion guided supervision aids domain alignment in the noise space and enhances restoration quality and generalization.

\subsubsection{Diffusion-Guided Noise Space Adaptation}
Noise-DA works directly in the diffusion noise space, as opposed to conventional domain adaptation approaches which are conducted in either pixel or feature space. Suppose that $x_s$ and $x_r$ represent synthetic and real degraded inputs, respectively. In both cases, the inputs are fed into a common restoration network $R(\cdot)$ that produces restored outputs:
\begin{equation}
\begin{aligned}
y_s &= R(x_s), y_r = R(x_r).
\end{aligned}
\label{eq:restoration}
\end{equation}
The reconstructed images are then used as conditioning inputs for a diffusion model which predicts the resulting noise distributions. Noise-DA reduces noise representation difference between prediction and target to bring synthetic noise and real noise to a noise space, thus enhancing the robustness of noise restoration and cross-domain generalization.

\subsubsection{Channel Shuffling and Residual Swapping Strategy}
\begin{figure}[!t]
    \centering
    \includegraphics[width=0.9\columnwidth]{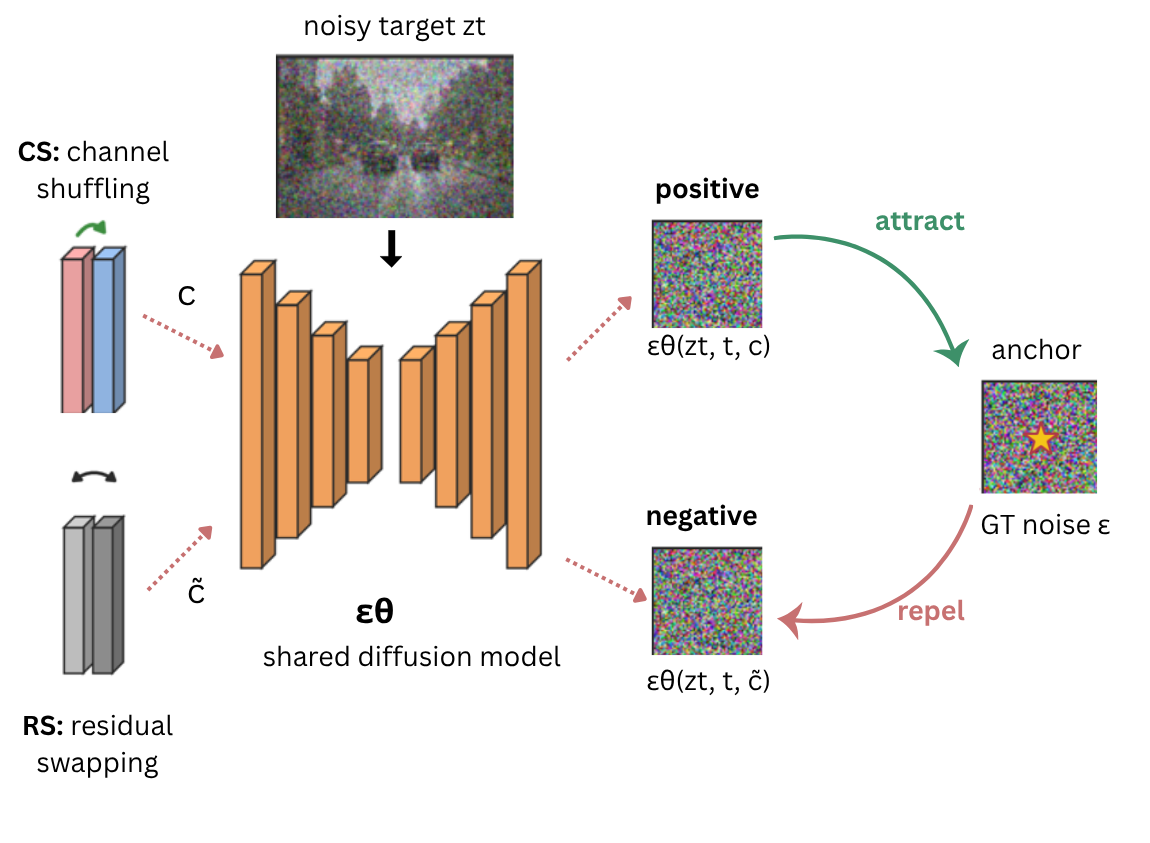}

    \caption{Channel Shuffling and Residual Swapping Mechanism in Noise-DA}
    \label{fig:fig7}
\end{figure}

For boosting domain generalization, Noise-DA introduces shuffle channels \cite{46Liao2025DenoisingAsAdaptation} and residual swap strategies during training, where the shuffle channels and the residual swaps are shown in Fig.~\ref{fig:fig7}. These strategies engage in the learning of representations that are not specific to the domain and shift the reliance of the domain-specific features.

The decomposition of the conditioning information in the diffusion model by shuffling the channels encourages degradation of information and not the appearance patterns specific to the domain. Residual swapping can exchange residual information among the training samples and form positive and negative noise pairs to enhance differentiation between the representations of degradation and increase domain alignment. Combined, these mechanisms enhance restoration robustness and help achieve the better cross-domain generalization performance of Noise-DA in various real-world degradation conditions.

\subsubsection{Multi-Source Training Strategy}
A multi-source training strategy was chosen to enhance the robustness of the restoration model and generalization. Several autonomous driving and image restoration datasets were merged and presented to the model with various degradation distributions such as weather, illumination, and sensor degradations. The restoration network is trained with a wide variety of degradation types, learning more general representations and better performance on the out-of-sample real-world scenarios.

Noise-DA combines domain adaptation with multi-source training to learn representations of the degradation that are more robust to a variety of environmental conditions. This enhanced robustness is especially helpful in applications for autonomous driving where image restoration has an impact on the subsequent image perception.

\subsection{Semantic Segmentation Framework}
\begin{figure}[!t]
    \centering

    \includegraphics[width=0.9\columnwidth]{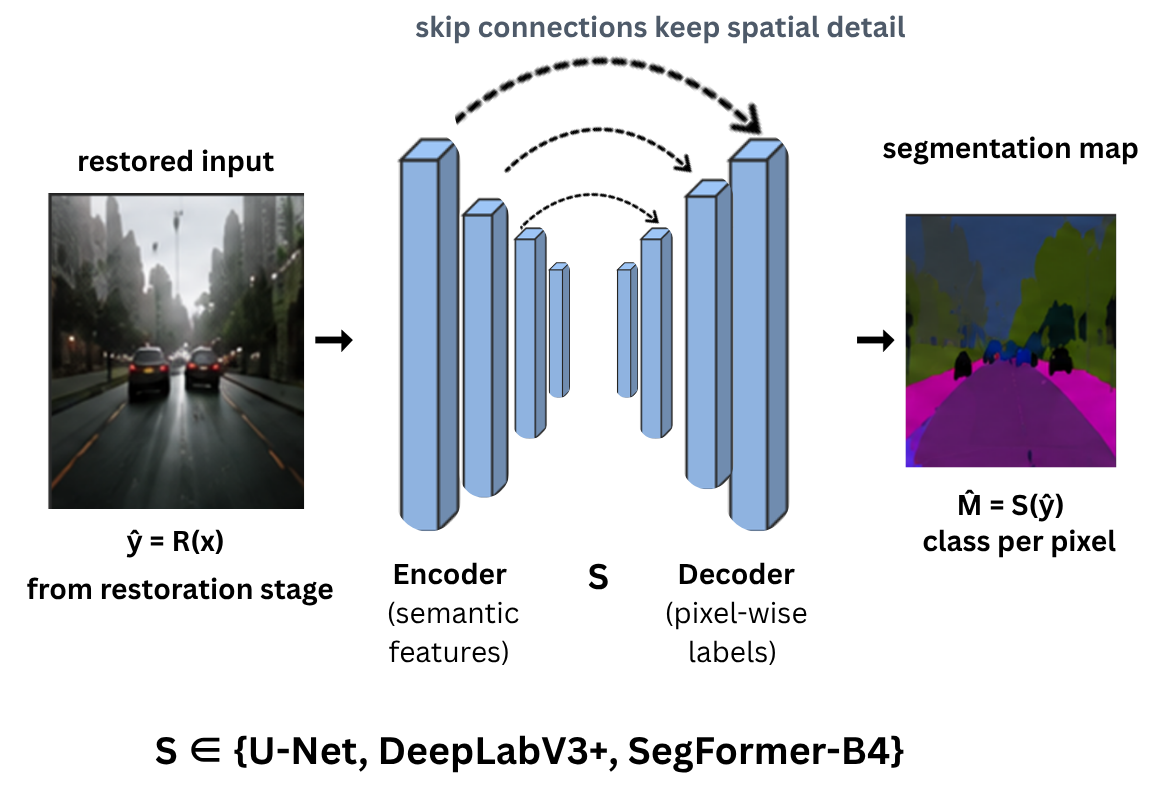}

    \caption{Semantic Segmentation Pipeline for Perception Evaluation}
    \label{fig:fig8}
\end{figure}

The performance of image restoration under adverse driving conditions was assessed by using semantic segmentation as the downstream perception task. Traditional image quality metrics are used to evaluate the restoration of images, but do not directly reflect the usefulness of the restored image in autonomous driving applications. Accordingly, semantic segmentation was used to evaluate the effect of denoising on scene understanding and accuracy of pixel-wise perception. Fig.~\ref{fig:fig8}. shows the semantic segmentation framework that was used to assess the influence of image restoration on perception. The restored images are fed to segmentation networks to obtain pixel-wise predictions for performance assessment under various degradation scenarios.

Different denoising methods were used to restore the images back and tested with the semantic segmentation models under various degradation conditions. Better segmentation performance shows that the restoration framework can maintain semantic information and recover visual structures that are crucial for autonomous driving perception.

To perform a thorough analysis, three different segmentation architectures were chosen: U-Net, DeepLabV3, and SegFormer, corresponding to different segmentation paradigms (convolution-based, context-aware, and transformer-based, respectively).

\subsubsection{Selected Segmentation Models}
Three popular semantic segmentation architectures were chosen to probe the downstream perception effects of image restoration, which belong to various segmentation paradigms. \\
\textbf{U-Net:} CNN-based encoder-decoder architecture with skip connections for efficient feature reconstruction and as a strong baseline segmentation model. \\
\textbf{DeepLabV3:} Context-aware Segmentation network using Atrous Spatial Pyramid Pooling (ASPP) to model multi-scale spatial information. \\
\textbf{SegFormer:} A transformer-based segmentation architecture that integrates hierarchical feature learning and context modeling across the whole image, with good robustness and generalization ability.

\subsubsection{Perception-Oriented Evaluation Strategy}
This work takes a perception-oriented evaluation strategy instead of using only the image quality metrics that are commonly used in conventional restoration studies. After being restored using various denoising models, images are then used as input to semantic segmentation networks to evaluate the influence of these restored images on the downstream scene understanding task.

The segmentation performance is compared to degraded and restored images for each approach to restoration. The results of improvements in both IoU and mIoU show that the restoration framework can restore semantic information and improve perception performance in the difficult driving scenarios. This type of evaluation gives more real-world assessment to the effectiveness of restoration than image quality measures.

\section{EXPERIMENTAL SETUP}
This section presents the datasets, degradation categories, model configurations, and training settings used to evaluate the proposed UDA-Restoration framework under diverse autonomous driving conditions.

\subsection{Dataset}
To evaluate the experimental results, six publicly available datasets of common research applications in the fields of autonomous driving, image restoration and adverse weather perception were explained in Table ~\ref{tab:tab3}. Different driving scenes and environmental conditions, weather effects and sensor degradations are provided in these datasets.

\begin{table}[!t]
\caption{Datasets Used in This Study}
\label{tab:tab3}
\centering
\footnotesize
\begin{tabular}{p{1.8cm} p{2.0cm} p{3.7cm}}
\toprule
\textbf{Dataset} & \textbf{Purpose} & \textbf{Characteristics} \\
\midrule

Cityscapes &
Segmentation \& Restoration &
Urban driving scenes with pixel-level annotations \\

ACDC &
Segmentation \& Adverse Weather &
Fog, rain, snow, and nighttime driving conditions \\

KITTI Noisy &
Restoration &
Real driving scenes with sensor and environmental noise \\

RainDAR &
Restoration &
Rain-affected driving sequences \\

UA-DETRAC &
Restoration &
Traffic surveillance under varying weather conditions \\

BDK &
Restoration &
Real-world noisy and degraded driving scenes \\

\bottomrule
\end{tabular}
\end{table}

Cityscapes and ACDC were chosen as the main sources for segmentation and adverse weather assessment, while KITTI Noisy, RainDAR, UA-DETRAC, and BDK were added to challenge the restoration models with various distributions of adverse weather and enhance cross-domain generalization.

\subsection{Image Degradation Generation}
Several different degradation types were considered when training and evaluating the autonomous driving system to simulate realistic conditions. These degradations are summarized in Table.~\ref{tab:tab4} are typical of those faced by a driver in real world driving conditions due to environmental, weather, illumination, motion and sensor factors.

\begin{table}[!t]
\caption{Degradation Types Used in the Experiments}
\label{tab:tab4}
\centering
\footnotesize
\setlength{\tabcolsep}{3pt}
\begin{tabularx}{\columnwidth}{p{0.28\columnwidth} X}
\toprule
\textbf{Category} & \textbf{Degradation Types} \\
\midrule

Weather-Based &
Rain, Fog, Snow \\

Noise-Based &
Gaussian Noise, Sensor Noise \\

Motion-Based &
Motion Blur \\

Illumination-Based &
Low-Light, Illumination Variation, Sun Glare \\

Sensor-Based &
Sensor Distortion, Vignette \\

Environmental &
Dust \\

\bottomrule
\end{tabularx}
\end{table}

\begin{figure}[!t]
    \centering
    \includegraphics[width=0.9\columnwidth]{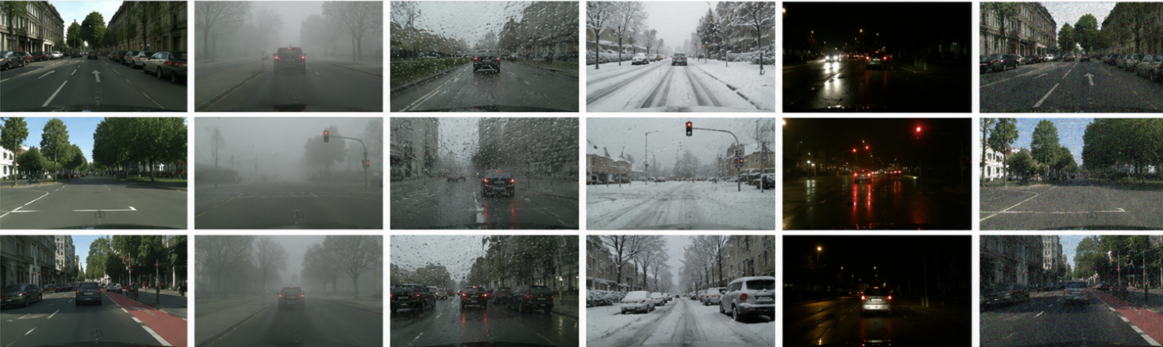}

    \caption{Examples of Driving Scene Degradations}
    \label{fig:fig9}
\end{figure}

Fig.~\ref{fig:fig9} illustrates representative weather, illumination, motion, and sensor-induced degradations encountered in autonomous driving environments.

\subsection{Restoration Model Configuration}
In the paired denoising experiments, DnCNN and NAFNet were trained with paired noisy-clean images samples, which were synthesized by applying synthetic degradation. The following methods were tested for denoising unpaired images: CycleGAN, Noise2Void-NAFNet, Blind2Unblind-UNet, NAFNet-SCPGANet and Noise-DA. The primary restoration framework was determined to be Noise-DA because of its ability to achieve better restoration results and cross-domain generalization.

\subsection{Segmentation Model Configuration}
To study the effect of image restoration on the perception of the downstream task, 3 representative architectures were used: CNN-based, context-aware, and transformer-based segmentation architectures, including U-Net, DeepLabV3, and SegFormer were selected. The data sets were split equally, with the same training and test sets for all models, to make a fair comparison between restoration methods.

\subsection{Evaluation Metrics}
For testing the proposed UDA-Restoration framework, Table ~\ref{tab:tab5}
 shows the image restoration and semantic segmentation metrics. Restoration metrics are used to evaluate the quality of reconstruction and the preservation of structure while segmentation metrics are used to evaluate the effect of restoration on the perception performance in the downstream.

\begin{table}[!t]
\caption{Evaluation Metrics}
\label{tab:tab5}
\centering
\footnotesize
\setlength{\tabcolsep}{3pt}
\begin{tabularx}{\columnwidth}{p{0.34\columnwidth} X}
\toprule
\textbf{Task} & \textbf{Metrics} \\
\midrule

Image Restoration &
PSNR, SSIM, MAE, MSE \\

Semantic Segmentation &
IoU, mIoU, Dice Loss \\

\bottomrule
\end{tabularx}
\end{table}

The fidelity of restoration and accuracy of reconstruction were assessed using PSNR, SSIM, MAE and MSE. IoU, mIoU and Dice Loss were used for semantic segmentation to measure class-wise and overall perception performance. These metrics, combined, offer a full picture of the image restoration quality and downstream semantic understanding.

\subsection{Implementation Details}
The experiments were all performed on a workstation with a 16 GB NVIDIA GPU and are implemented using deep learning frameworks in Python. The models were trained with Adam optimizer and validation-based checkpointing and early stopping techniques were applied to ensure stable training for the restoration and semantic segmentation models.

The final noisy-clean image pair denoising dataset comprised 7550 clean images and 17001 noisy images from several public datasets, comprising a total of about 30 GB of training data. Images were preprocessed using standard procedures before training, such as resizing and normalizing images.

The final Noise-DA model was trained for 100 epochs using a batch size of 4, and initial learning rate of $10^{-4}$. The training took around four days of computation on the available hardware with a large-scale dataset and the computational complexity of diffusion-based restoration. For restoration and semantic segmentation evaluation, the best performing checkpoints were then used.

\section{RESULTS AND DISCUSSION}
This section presents and discusses the experimental findings of the proposed framework from both restoration and perception perspectives. The analysis first compares quantitative image restoration performance across the evaluated methods, followed by qualitative assessment under challenging degradation conditions and evaluation of the impact of restoration on downstream semantic segmentation.

\subsection{Quantitative Image Restoration Performance}
The effectiveness of the restoration process under various types of degradation was assessed by comparing the supervised, self-supervised and unsupervised restoration methods with PSNR and SSIM. They show how well the structure and image can be reconstructed as higher values represent better reconstruction, fidelity and structural preservation.

\begin{table}[!t]
\caption{Quantitative Comparison of Image Restoration Methods}
\label{tab:tab6}
\centering
\scriptsize
\begin{tabular}{p{2.2cm} p{2.6cm} c c}
\toprule
\textbf{Method} & \textbf{Learning Paradigm} & \textbf{PSNR (dB) $\uparrow$} & \textbf{SSIM $\uparrow$} \\
\midrule

CycleGAN &
Unpaired GAN &
25.84 &
0.7912 \\

Blind2Unblind UNet &
Self-Supervised &
28.74 &
0.8644 \\

NAFNet Unpaired &
Self-Supervised &
29.15 &
0.8814 \\

N2V-NAFNet &
Noise2Void &
29.34 &
0.8875 \\

SURE-NAFNet &
SURE Based &
29.88 &
0.8921 \\

N2N-NAFNet &
Noise2Noise &
30.11 &
0.9012 \\

NAFNet + SCPGABNet &
Attention-Guided &
31.04 &
0.9142 \\

\textbf{Noise-DA (Proposed)} &
\textbf{UDA + Diffusion Guidance} &
\textbf{31.63} &
\textbf{0.9777} \\

\bottomrule
\end{tabular}
\end{table}

Table ~\ref{tab:tab6} demonstrates the progressive improvement in restoration results from conventional GAN-based approaches to self-supervised and diffusion-guided methods. Self-supervised methods showed significant improvements in both PSNR and SSIM with CycleGAN being the least successful. N2N-NAFNet and NAFNet-SCPGABNet showed promising restoration performance with a PSNR greater than 30 dB among these methods.

LPIPS and FID were further calculated on pairs of clean and denoised images for a further evaluation of perceptual restoration quality. Under the proposed Noise-DA scheme, the LPIPS score is 0.0123 ± 0.0025 and the FID score is 6.2490 which means high perceptual similarity and good distributional consistency between the restored and reference images. The results in this paper reinforce the capabilities of Noise-DA in maintaining visual structures and reduce perceptual distortion.

The proposed framework Noise-DA achieved 31.63 dB PSNR and 0.9777 SSIM, representing the best performance among the unsupervised methods and matching the strongest supervised result. The diffusion-guided domain adaptation strategy adopted by Noise-DA is key to achieving its superior performance, allowing for effective alignment of the synthetic and real degradation distributions while maintaining the structural image information.

The Noise-DA convergence behaviour is shown in Fig.~\ref{fig:fig10}. The restoration loss gradually improves during the training process, whereas SSIM continuously improves, showing robust optimization and learning of degradation-invariant representations.

These findings show that the diffusion guided unsupervised domain adaptation can achieve a performance on restoration comparable to supervised approaches without the need of paired clean–degraded training data.

\subsection{Comparison of Supervised and Unsupervised Restoration Methods}
The proposed Noise-DA framework was compared to the supervised restoration models to assess the performance of unsupervised domain adaptation. The goal was to see if a method could be trained to restore an image using diffusion-guided adaptation that was like one trained using pairs of clean and degraded images. Table ~\ref{tab:tab7} presents the quantitative comparison between supervised and unsupervised restoration approaches.

\begin{table}[!t]
\caption{Comparison of Supervised and Unsupervised Restoration Methods}
\label{tab:tab7}
\centering
\scriptsize
\setlength{\tabcolsep}{2.5pt}
\begin{tabularx}{\columnwidth}{p{0.23\columnwidth} X p{0.16\columnwidth} p{0.12\columnwidth}}
\toprule
\textbf{Model} & \textbf{Training Data} & \textbf{PSNR (dB) $\uparrow$} & \textbf{SSIM $\uparrow$} \\
\midrule

DnCNN-B &
Cityscapes &
21.85 &
0.6367 \\

DnCNN-S &
ACDC &
12.44 &
0.4783 \\

NAFNet &
Cityscapes &
30.95 &
0.9005 \\

NAFNet (Combined) &
Cityscapes + ACDC &
31.63 &
0.9777 \\

\textbf{Noise-DA (Proposed)} &
\textbf{Multi-Source Unsupervised} &
\textbf{31.63} &
\textbf{0.9777} \\

\bottomrule
\end{tabularx}
\end{table}

Under adverse conditions, DnCNN showed limited generalization while performance of restoration performance was significantly improved in NAFNet as demonstrated in Table ~\ref{tab:tab7}.The proposed Noise-DA framework achieved 31.63 dB PSNR and 0.9777 SSIM, matching the best supervised NAFNet result while requiring no paired target-domain supervision.

The results prove that diffusion-guided domain adaptation can successfully close the gap between supervised and unsupervised restoration and serve as a practical approach for the real-world autonomous driving scenario where paired clean references are unavailable.

\begin{figure}[!t]
    \centering

    \includegraphics[width=0.9\columnwidth]{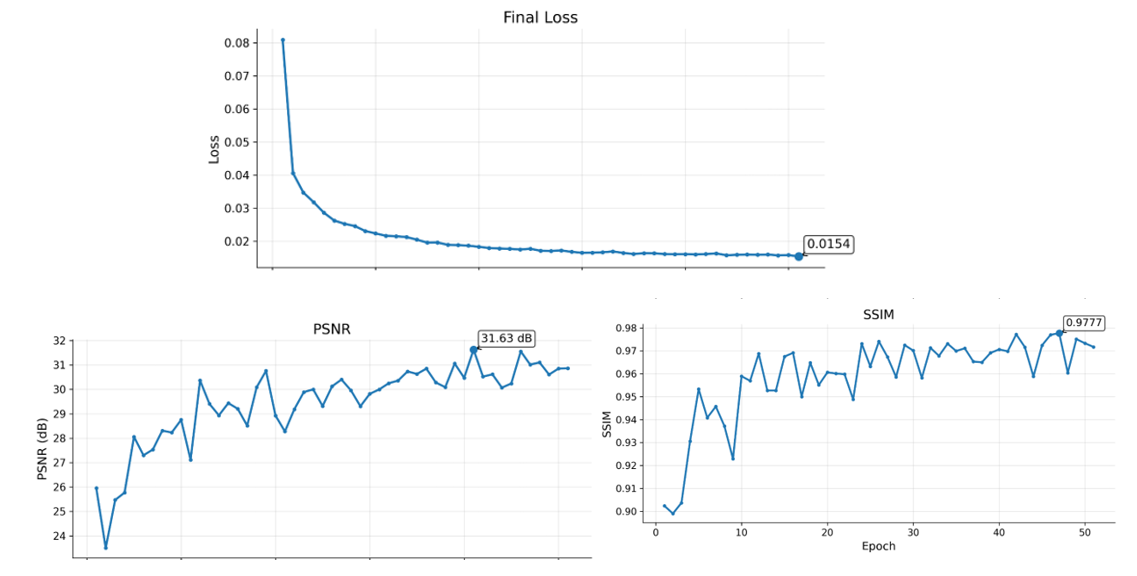}

    \caption{Convergence training within the framework of the proposed Noise-DA}
    \label{fig:fig10}
\end{figure}

Fig.~\ref{fig:fig10}. illustrates the convergence behavior of the proposed Noise-DA framework during training. The restoration loss decreases rapidly in the initial epoch and gradually converges to a stable value of 0.0154, indicating effective optimization and successful minimization of reconstruction error. Concurrently, the PSNR increases steadily and reaches 31.63 dB, demonstrating the model's enhanced capability to suppress noise and reconstruct high-fidelity images. Similarly, the SSIM improves throughout the training process, attaining a value of 0.9777, which reflects excellent preservation of structural information, texture details, and perceptual image quality. The simultaneous reduction in loss and improvement in both PSNR and SSIM confirms the stable convergence, robustness, and effectiveness of the proposed Noise-DA framework for image restoration tasks.

\subsection{Qualitative Restoration Analysis}
Qualitative evaluation, in addition to quantitative evaluation, were also conducted to analyze the visual quality and generalization ability of the restoration methods in realistic situations.

\begin{figure}[!t]
    \centering

    \includegraphics[width=0.9\columnwidth]{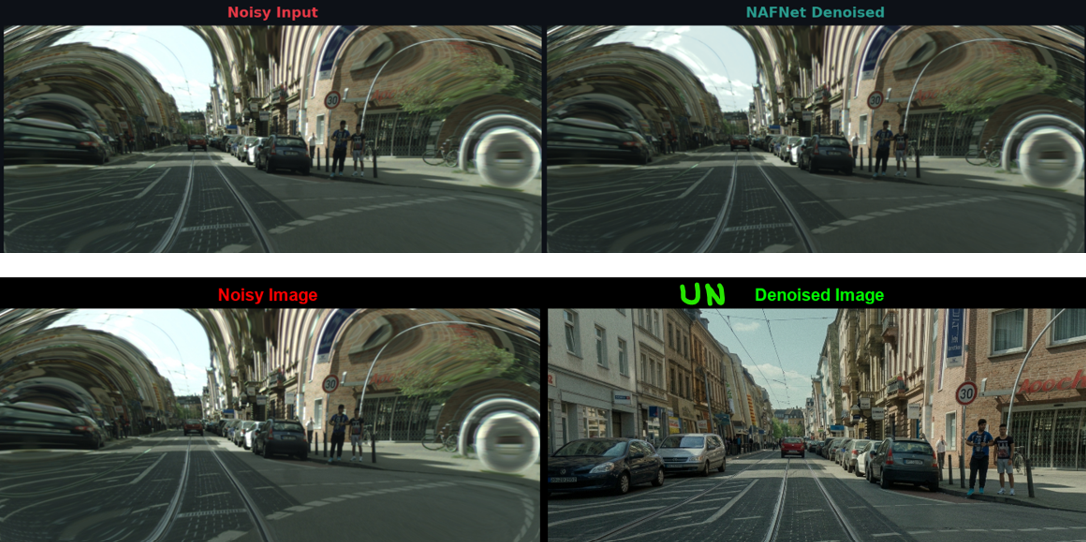}

    \caption{Qualitative comparison of Sensor distortion scenes}
    \label{fig:fig11}
\end{figure}

The results are shown in Fig.~\ref{fig:fig11}. for severe sensor distortion. Some distortion artifacts are reduced using NAFNet, but the remaining artifacts are noted as structural degradation. However, Noise-DA can recover the geometry of the road, boundaries of objects, and details from the scene, with a better visual consistency and visual perception.

The overall conclusions drawn from the qualitative results are in line with those obtained from quantitative results, and they show that diffusion-guided domain adaptation achieves better restoration and generalization across domains under different environmental degradations.

\subsection{Impact of Denoising on Semantic Segmentation Performance}
The efficiency of image restoration algorithms for downstream perception was evaluated on degraded and restored driving images by applying U-Net, DeepLabV3+ and SegFormer. Semantic segmentation offers a task-oriented measure of restoration quality unlike that of PSNR and SSIM, by measuring the performance of scene understanding.

Table ~\ref{tab:tab8} summarizes the segmentation results. The image degradations greatly affected accuracy of segmentation, while the restored images enhanced both the visibility of objects and the delineation of boundaries, as well as the overall interpretation of the scene. On the tested architectures, SegFormer outperforms other architectures with 62.19\% mIoU, showing that SegFormer is robust under difficult driving situations.

\begin{table}[!t]
\caption{Semantic Segmentation Performance Comparison}
\label{tab:tab8}
\centering
\scriptsize
\begin{tabular}{p{2.8cm} c c}
\toprule
\textbf{Model} & \textbf{Dice Score} & \textbf{mIoU} \\
\midrule

U-Net (Clean) &
0.6997 &
0.6486 \\

U-Net (Blur) &
0.5436 &
0.4967 \\

U-Net (Fog) &
0.5889 &
0.5260 \\

Attention U-Net &
0.6821 &
0.6302 \\

DeepLabV3+ &
0.4548 &
0.4103 \\

SegFormer-B4 &
-- &
\textbf{0.6219} \\

\bottomrule
\end{tabular}
\end{table}

\begin{figure}[!t]
    \centering
    \includegraphics[width=0.9\columnwidth]{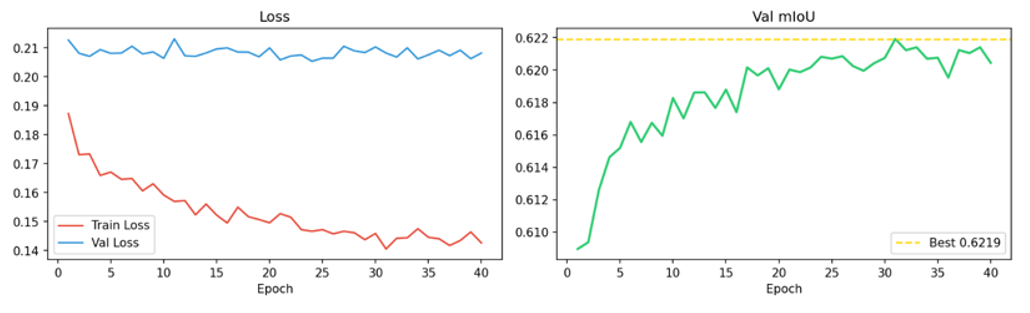}

    \caption{Segformer Training Curves}
    \label{fig:fig12}
\end{figure}

Fig.~\ref{fig:fig12}. shows the training convergence of SegFormer. The training loss decreases steadily from 0.188 to 0.143, indicating effective optimization of the segmentation objective. Meanwhile, the validation mIoU increases progressively and reaches a peak value of 0.6219 before stabilizing, demonstrating improved segmentation accuracy and stable model convergence without noticeable performance degradation.

\begin{figure}[!t]
    \centering
    \includegraphics[width=0.9\columnwidth]{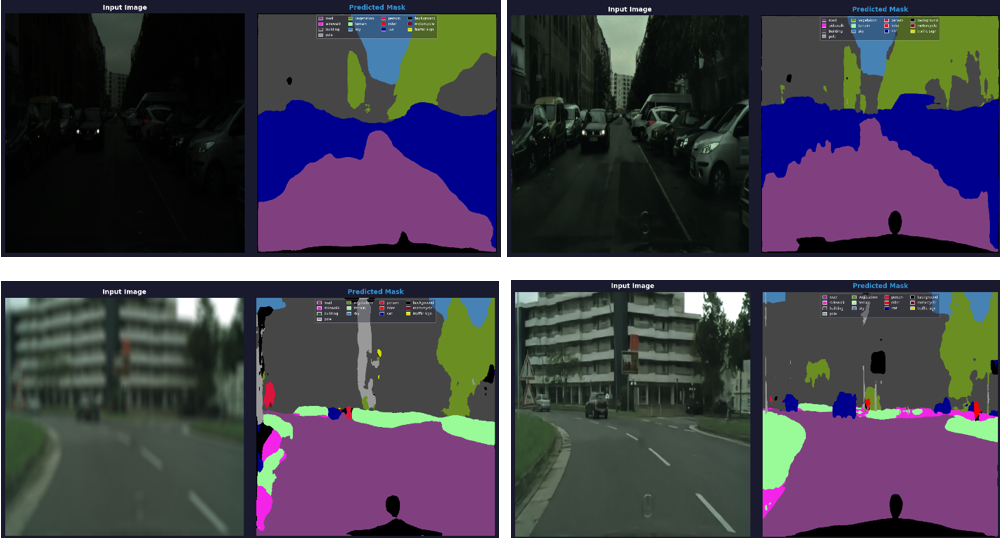}

    \caption{Qualitative Impact of Image Restoration on Semantic Segmentation Performance}
    \label{fig:fig13}
\end{figure}
Fig.~\ref{fig:fig13}. shows the segmentation results using qualitative segmentation on degraded image and restored image. The restoration enhances the visibility of the scene, the edges of the objects, and semantic consistency, leading to more precise segmentation results.

\begin{figure}[!t]
    \centering
    \includegraphics[width=0.7\columnwidth]{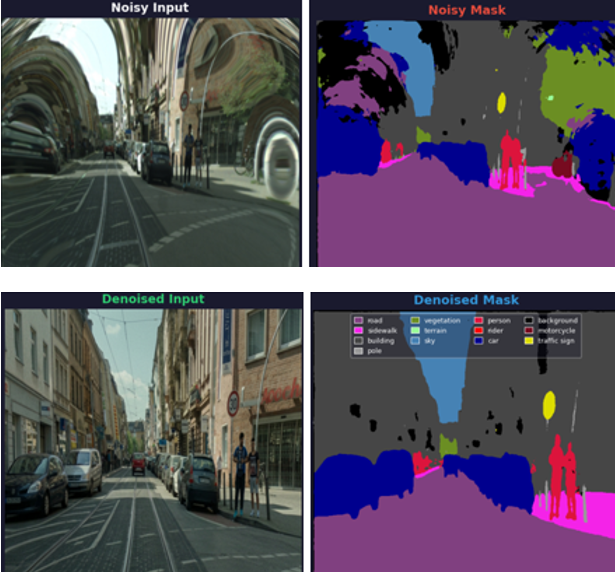}

    \caption{Comparison of Segmentation on Supervised vs Unsupervised Denoised image}
    \label{fig:fig14}
\end{figure}
The results of supervised (NAFNet) and unsupervised (Noise-DA) restoration is compared in Fig.~\ref{fig:fig14}. Both methods yield better perception results, but Noise-DA provides better segmentation boundaries and more coherent semantic regions in severe degradations, showing its better cross-domain generalization.

Overall, the findings indicate that image restoration is effective not only for image restoration visual quality but also in improving downstream perception performance, and that the proposed noise-DA is effective in almost all challenging autonomous driving situations.

\subsection{Discussion}
The experimental results show that image restoration can effectively enhance the performance of perception in adverse weather conditions, light changes, and degradations caused by sensors in autonomous driving scenarios. Supervised approaches like DnCNN and NAFNet perform well on paired data but fail to generalize well when faced with unseen degradation.

The proposed Noise-DA framework, on the other hand, was consistently able to obtain the best restoration results by performing noise space adaptation using diffusion guidance. Noise-DA did not need paired domain supervision and learned degradation-invariant representations that were robust against synthetic noise, achieving the best restoration results of 31.63 dB PSNR and 0.9777 SSIM.

A further validation of the benefits of restoration was obtained through semantic segmentation experiments, demonstrating that restored images led to better scene understanding than degraded ones. Of the tested techniques, Noise-DA yielded the best segmentation results, indicating that restoration quality correlates well with the perception results in the downstream. 

Moreover, the multi-source training strategy significantly helped in boosting cross-domain generalization by training the model with different types of weather and degradation. This allowed successful restoration in complex real-world situations, such as in snow, low light conditions, and sensor distortions.

In general, the results validate that diffusion-guided unsupervised domain adaptation is a promising and scalable approach for robust image restoration and perception enhancement in autonomous driving systems.

\section{CONCLUSION}
This paper introduced a domain adaptation method for image restoration in autonomous driving environment, which is unsupervised diffusion-guided (UDA-Restoration). The proposed approach provides effective and flexible adaptation between synthetic and real-world degradations by leveraging multi-source degradation learning and Noise-DA.
Experimental results showed that Noise-DA achieved 31.63 dB PSNR and 0.9777 SSIM, outperforming the evaluated self-supervised and unsupervised baselines while matching the best supervised restoration result. Qualitative analysis also validated its performance of recovering visual details and scene structures in challenging scenarios, such as snow, low-light conditions, and sensor distortions. The effectiveness of restoration was also confirmed by semantic segmentation experiments with U-Net, DeepLabV3, and SegFormer. The enhanced image quality resulted in better segmentation performance on restored images, showing that improved restoration can contribute directly to more reliable scene understanding. Overall, the results indicate that diffusion-guided unsupervised domain adaptation provides a robust and scalable solution for image restoration in autonomous driving perception systems. Future evaluation may extend the framework to additional perception tasks and larger-scale real-world driving scenarios.

\nocite{*}
\bibliographystyle{IEEEtran}
\bibliography{references}

@article{1Li2025WControlUDA,
  author  = {Li, J. and others},
  title   = {W-ControlUDA: Weather-Controllable Diffusion-Assisted Unsupervised Domain Adaptation for Semantic Segmentation},
  journal = {IEEE Transactions on Pattern Analysis and Machine Intelligence},
  volume  = {47},
  number  = {3},
  pages   = {1824--1836},
  year    = {2025},
  month   = {Mar},
  doi     = {10.1009/TPAMI.2025.10900417}
}

@article{2Jung2025CrossAttention,
  author  = {Jung, H. and others},
  title   = {Cross-Attention Joint Denoising and Semantic Segmentation for Robust Image Analysis},
  journal = {IEEE Transactions on Image Processing},
  volume  = {34},
  pages   = {2411--2424},
  year    = {2025},
  doi     = {10.1109/TIP.2024.11284809}
}

@inproceedings{3Pan2025WeatherAware,
  author    = {Pan, J. and others},
  title     = {Exploring Weather-Aware Aggregation and Adaptation for Semantic Segmentation under Adverse Conditions},
  booktitle = {Proceedings of the IEEE/CVF International Conference on Computer Vision (ICCV)},
  pages      = {5422--5432},
  year       = {2025}
}

@inproceedings{4Truong2024CONDA,
  author    = {Truong, T. and others},
  title     = {CONDA: Continual Unsupervised Domain Adaptation Learning in Visual Perception for Autonomous Driving},
  booktitle = {Proceedings of the IEEE/CVF Conference on Computer Vision and Pattern Recognition Workshops (CVPRW)},
  pages      = {3104--3114},
  year       = {2024}
}

@article{5Ibrahum2026AdversarialML,
  author  = {Ibrahum, M. and others},
  title   = {Adversarial Machine Learning Research: Modeling Attack Vectors and Developing Robust Defense Strategies for AI Systems},
  journal = {Artificial Intelligence Review},
  volume  = {59},
  number  = {1},
  pages   = {148},
  year    = {2026},
  month   = {Jan},
  doi     = {10.1007/s10462-024-11014-8}
}

@inproceedings{6Yu2025ADLD,
  author    = {Yu, J. and Sun, Z. and Chi, C. B. and Yang, J. and Luo, S.},
  title     = {Adversarially Domain-Adaptive Latent Diffusion for Unsupervised Semantic Segmentation},
  booktitle = {Proceedings of the IEEE/CVF Conference on Computer Vision and Pattern Recognition Workshops (CVPRW)},
  pages      = {2615--2624},
  year       = {2025}
}

@inproceedings{7Liao2025DenoisingAdaptation,
  author    = {Liao, K. and Yue, Z. and Wang, Z. and Loy, C. C.},
  title     = {Denoising as Adaptation: Noise-Space Domain Adaptation for Image Restoration},
  booktitle = {International Conference on Learning Representations (ICLR)},
  pages      = {1--15},
  year       = {2025},
  month      = may
}

@inproceedings{8Ye2022NighttimeTracking,
  author    = {Ye, J. and Fu, C. and Zheng, G. and Paudel, D. P. and Chen, G.},
  title     = {Unsupervised Domain Adaptation for Nighttime Aerial Tracking},
  booktitle = {Proceedings of the IEEE/CVF Conference on Computer Vision and Pattern Recognition (CVPR)},
  pages      = {8886--8895},
  year       = {2022},
  month      = jun
}

@article{9Wei2024NighttimeObjectTracking,
  author  = {Wei, H. and others},
  title   = {Unsupervised Nighttime Object Tracking Based on Transformer and Domain Adaptation Fusion Network},
  journal = {IEEE Access},
  volume  = {12},
  pages   = {43117--43128},
  year    = {2024},
  month   = mar,
  doi     = {10.1109/ACCESS.2024.3378117}
}

@article{10Peng2025DomainAdaptiveDiffusion,
  author  = {Peng, H. and Wang, L. and Zhou, G.},
  title   = {Domain-Adaptive Diffusion: Progressive Distribution Alignment for Unsupervised Domain Adaptation},
  journal = {IEEE Transactions on Pattern Analysis and Machine Intelligence},
  volume  = {47},
  number  = {5},
  pages   = {3120--3133},
  year    = {2025},
  month   = may,
  doi     = {10.1109/TPAMI.2025.10421190}
}

@article{11Ridley2025ControllingSemantics,
  author  = {Ridley, L. and Kapoor, S. and Vance, A.},
  title   = {Controlling Semantics of Diffusion-Augmented Data for Unsupervised Domain Adaptation},
  journal = {IET Computer Vision},
  volume  = {19},
  number  = {2},
  pages   = {142--155},
  year    = {2025},
  month   = feb,
  doi     = {10.1049/cvi2.12311}
}

@article{12Benbarka2025SeeingThroughWeather,
  author  = {Benbarka, E. and El-Sallab, M. and Gadallah, Y.},
  title   = {Seeing Through the Weather: Controlled Generative Data Augmentation for Robust Autonomous Vehicle Perception},
  journal = {IEEE Transactions on Intelligent Transportation Systems},
  volume  = {26},
  number  = {4},
  pages   = {4810--4823},
  year    = {2025},
  month   = apr,
  doi     = {10.1109/TITS.2025.3409112}
}

@inproceedings{13Li2024WeatherGeneralSpecific,
  author    = {Li, Y. and others},
  title     = {Learning Weather-General and Weather-Specific Features for Image Restoration Under Multiple Adverse Weather Conditions},
  booktitle = {Proceedings of the IEEE/CVF Conference on Computer Vision and Pattern Recognition (CVPR)},
  pages      = {2931--2941},
  year       = {2024}
}

@article{14An2025MoEDriven,
  author  = {An, X. and Zhu, M. and Wang, Y.},
  title   = {An MoE-Driven Unified Image Restoration Framework for Adverse Weather Conditions},
  journal = {IEEE Transactions on Circuits and Systems for Video Technology},
  volume  = {35},
  number  = {2},
  pages   = {840--853},
  year    = {2025},
  month   = feb,
  doi     = {10.1109/TCSVT.2024.10801389}
}

@inproceedings{15Wang2025MOERL,
  author    = {Wang, H. and Zhang, J. and Fu, X.},
  title     = {MOERL: When Mixture-of-Experts Meet Reinforcement Learning for Adverse Weather Image Restoration},
  booktitle = {Proceedings of the IEEE/CVF International Conference on Computer Vision (ICCV)},
  pages      = {2210--2220},
  year       = {2025}
}

@article{16Lin2025MultiComponent,
  author  = {Lin, Z. and Fu, X. and Huang, J.},
  title   = {Multi-Component Decomposition Strategy for Mixed-Weather Image Restoration},
  journal = {IEEE Signal Processing Letters},
  volume  = {32},
  pages   = {541--545},
  year    = {2025},
  doi     = {10.1109/LSP.2025.3401140}
}

@article{17Fang2025WeatherCycle,
  author  = {Fang, J. and Guo, R. and Zhao, X.},
  title   = {WeatherCycle: Fully Unsupervised Multi-Weather Image Restoration via Color-Space Decoupling},
  journal = {IEEE Transactions on Multimedia},
  volume  = {27},
  pages   = {1910--1922},
  year    = {2025},
  doi     = {10.1109/TMM.2024.3395112}
}

@article{18Liu2025WeatherNormalization,
  author  = {Liu, D. and Pan, J. and Sun, R.},
  title   = {Decoupling Weather Normalization and Semantic Understanding for Robust Autonomous Driving Perceptions},
  journal = {IEEE Robotics and Automation Letters},
  volume  = {10},
  number  = {3},
  pages   = {1450--1457},
  year    = {2025},
  month   = mar,
  doi     = {10.1109/LRA.2025.3408894}
}

@article{19Wang2025LaplacianPyramid,
  author  = {Wang, Y. and Zhang, J. and Sigal, L.},
  title   = {Unified Laplacian Pyramid Enhancement and Domain Adaptation for Low-Light Object Detection},
  journal = {IEEE Transactions on Image Processing},
  volume  = {34},
  pages   = {1102--1115},
  year    = {2025},
  doi     = {10.1109/TIP.2024.3384561}
}

@article{20Wang2025DiffAD,
  author  = {Wang, X. and others},
  title   = {DiffAD: A Unified Diffusion Modeling Approach for Autonomous Driving},
  journal = {arXiv preprint arXiv:2503.12170},
  year    = {2025}
}

@article{21Wang2025TaskDrivenEnhancement,
  author  = {Wang, S. and Liang, Y. and Zhang, T.},
  title   = {Task-Driven Image Enhancement Networks for Autonomous Driving Scene Parsing},
  journal = {IEEE Transactions on Intelligent Transportation Systems},
  volume  = {26},
  number  = {1},
  pages   = {320--333},
  year    = {2025},
  month   = jan,
  doi     = {10.1109/TITS.2024.3361102}
}

@article{22Pan2025ObjectAwareUDA,
  author  = {Pan, X. and Wei, H. and Fang, J.},
  title   = {Object-Aware Unsupervised Domain Adaptation for Nighttime Perception Refinement},
  journal = {IEEE Transactions on Intelligent Vehicles},
  volume  = {10},
  number  = {2},
  pages   = {290--302},
  year    = {2025},
  month   = apr,
  doi     = {10.1109/TIV.2024.10801389}
}

@inproceedings{23Guan2025WeatherBench,
  author    = {Guan, A. and Yu, J. and Li, H.},
  title     = {WeatherBench: A Real-World Benchmark Dataset for All-in-One Adverse Weather Image Restoration},
  booktitle = {Proceedings of the ACM International Conference on Multimedia (ACM MM)},
  pages      = {1205--1215},
  year       = {2025}
}

@article{24Wen2025AllInOneWeather,
  author  = {Wen, Y. and Gao, T. and Li, Z. and Zhang, J. and Zhang, K. and Chen, T.},
  title   = {All-in-One Weather-Degraded Image Restoration Via Adaptive Degradation-Aware Self-Prompting Model},
  journal = {IEEE Transactions on Multimedia},
  volume  = {27},
  pages   = {3343--3355},
  year    = {2025}
}

@article{25Cui2025MultiScaleFeature,
  author  = {Cui, Y. and Zhu, J. and Knoll, A.},
  title   = {Enhancing Perception for Autonomous Vehicles: A Multi-Scale Feature Modulation Network for Image Restoration},
  journal = {IEEE Transactions on Intelligent Transportation Systems},
  volume  = {26},
  number  = {4},
  pages   = {4621--4633},
  year    = {2025},
  month   = apr
}

@article{26Zheng2023RobustPerception,
  author  = {Zheng, Z. and Cheng, Y. and Xin, Zhichao and Yu, Zhibin and Zheng, B.},
  title   = {Robust Perception Under Adverse Conditions for Autonomous Driving Based on Data Augmentation},
  journal = {IEEE Transactions on Intelligent Transportation Systems},
  volume  = {24},
  number  = {12},
  pages   = {13916--13928},
  year    = {2023},
  month   = dec
}

@article{27Erkent2020SemanticSegmentation,
  author  = {Erkent, {\"O}. and Laugier, C.},
  title   = {Semantic Segmentation With Unsupervised Domain Adaptation Under Varying Weather Conditions for Autonomous Vehicles},
  journal = {IEEE Robotics and Automation Letters},
  volume  = {5},
  number  = {2},
  pages   = {3580--3587},
  year    = {2020},
  month   = apr
}

@inproceedings{28Pan2024DynamicSmallObjects,
  author    = {Pan, J. and Li, S. and Chen, Y. and Zhu, J. and Wang, L.},
  title     = {Towards Dynamic and Small Objects Refinement for Unsupervised Domain Adaptive Nighttime Semantic Segmentation},
  booktitle = {Proceedings of the IEEE/RSJ International Conference on Intelligent Robots and Systems (IROS)},
  pages      = {372--378},
  year       = {2024},
  month      = oct
}

@article{29Leng2026SnowyWeatherDenoising,
  author  = {Leng, W. and Huang, Y. and Zhang, X. and Hu, C.},
  title   = {Unsupervised Snowy-Weather Point Cloud Denoising via Two-Stage Filter-Network Collaboration},
  journal = {IEEE Transactions on Geoscience and Remote Sensing},
  volume  = {64},
  pages   = {1--13},
  year    = {2026},
  note    = {Art. no. 5701715}
}

@article{30Kumar2025VisionTechnologies,
  author  = {Kumar, H. and Mamoria, P. and Dewangan, D. K.},
  title   = {Vision Technologies in Autonomous Vehicles: Progress, Methodologies, and Key Challenges},
  journal = {International Journal of System Assurance Engineering and Management},
  volume  = {16},
  pages   = {4035--4068},
  year    = {2025}
}

@article{Zhang2023DeepLearningIQA,
  author  = {Zhang, X. and others},
  title   = {Deep Learning Based Image Quality Assessment: A Survey},
  journal = {Procedia Computer Science},
  volume   = {221},
  pages    = {1000--1005},
  year     = {2023},
  doi       = {10.1016/j.procs.2023.08.080}
}

@inproceedings{31Verma2025WeatherAffectedVision,
  author    = {Verma, A. and Warnekar, H. A. and Ghanekar, U.},
  title     = {Comprehensive Weather-Affected Automobile Vision Algorithm Based on SURF and Optical Flow},
  booktitle = {Proc. 3rd Int. Conf. Inventive Computing and Informatics (ICICI)},
  year       = {2025},
  doi        = {10.1109/ICICI65870.2025.11069973}
}

@article{43Hossain2025AdaptiveAerialRestoration,
  author  = {Hossain, M. Y. and Rakib, M. M. H. and Rajit, S. and Nijhum, I. R. and Rahman, R. M.},
  title   = {Adaptive and Automatic Aerial Image Restoration Pipeline Leveraging Pre-trained Image Restorer with Lightweight Fully Convolutional Network},
  journal = {Expert Systems with Applications},
  volume  = {259},
  pages   = {125210},
  year    = {2025},
  doi     = {10.1016/j.eswa.2024.125210}
}

@article{45Shi2026FIRSDE,
  author  = {Shi, X. and Xu, Z. and Ge, S. and Zhao, X. and Wang, X. and Zhang, W.},
  title   = {FIR-SDE: Fast Image Restoration via Mean-Reverting Stochastic Differential Equation},
  journal = {Complex \& Intelligent Systems},
  volume  = {12},
  pages   = {91},
  year    = {2026},
  doi     = {10.1007/s40747-025-01886-4}
}

@inproceedings{46Liao2025DenoisingAsAdaptation,
  author    = {Liao, Kang and Yue, Zongsheng and Wang, Zhouxia and Loy, Chen Change},
  title     = {Denoising as Adaptation: Noise-Space Domain Adaptation for Image Restoration},
  booktitle = {Proceedings of the International Conference on Learning Representations (ICLR)},
  year       = {2025}
}

@article{47Sakaridis2025ACDC,
  author  = {Sakaridis, C. and Wang, H. and Li, K. and Zurbr{\"u}gg, R. and Jadon, A. and Abbeloos, W.},
  title   = {ACDC: The Adverse Conditions Dataset With Correspondences for Robust Semantic Driving Scene Perception},
  journal = {IEEE Transactions on Pattern Analysis and Machine Intelligence},
  volume  = {48},
  number  = {3},
  year    = {2025}
}

@article{48Sowe2025PSPNetCityscapes,
  author  = {Sowe, E. A. and Sanyang, M. F. and Yahya, W. and Gegbe, H. G.},
  title   = {Semantic Segmentation in Self-Driving Cars Using Pyramid Parsing Network (PSPNet) on Cityscape Dataset},
  journal = {European Journal of Applied Science, Engineering and Technology},
  volume  = {3},
  number  = {1},
  year    = {2025},
  doi     = {10.59324/ejaset.2025.3(1).07}
}
\vspace{12pt}

\end{document}